# Spectral-Spatial Mamba for Hyperspectral Image Classification

Lingbo Huang [1], Yushi Chen [1]* and Xin He [1]

1 School of Electronics and Information Engineering, Harbin Institute of Technology, Harbin 150001, China; 22B905001@stu.hit.edu.cn(L.H.); hexin1@hit.edu.cn(X.H.)
* Correspondence: chenyushi@hit.edu.cn;

**Abstract:** Recently, Transformer has gradually attracted interest for its excellence in modeling the long-range dependencies of spatial-spectral features in HSI. However, Transformer has the problem of the quadratic computational complexity due to the self-attention mechanism, which is heavier than other models and thus has limited adoption in HSI processing. Fortunately, the recently emerging state space model-based Mamba shows great computational efficiency while achieving the modeling power of Transformers. Therefore, in this paper, we first proposed spectral-spatial Mamba (SS-Mamba) for HSI classification. Specifically, SS-Mamba mainly includes spectral-spatial token generation module and several stacked spectral-spatial Mamba blocks. Firstly, the token generation module converts any given HSI cube to spatial and spectral tokens as sequences. And then these tokens are sent to stacked spectral-spatial mamba blocks (SS-MB). Each SS-MB includes two basic mamba blocks and a spectral-spatial feature enhancement module. The spatial and spectral tokens are processed separately by the two basic mamba blocks, respectively. Besides, the feature enhancement module modulates spatial and spectral tokens using HSI sample's center region information. Therefore, the spectral and spatial tokens cooperate with each other and achieve information fusion within each block. The experimental results conducted on widely used HSI datasets reveal that the proposed SS-Mamba requires less processing time compared with Transformer. The Mamba-based method opens a new window for HSI classification. Code is available at https://github.com/mengduanjinghua/Spectral-spatial-Mamba-for-HSIC.

## 1. Introduction

Hyperspectral images (HSIs) can capture hundreds of narrow spectral bands across the electromagnetic spectrum, containing both rich spatial and spectral information [1]. Compared with traditional RGB images and multispectral images, HSIs can provide richer information about the land-covers, and thus have been widely exploited in many applications, such as environmental monitoring [2], precision agriculture [3], mineral exploration [4], and military reconnaissance [5]. Aiming at pixel-level category labeling, Classification is a fundamental task for HSI processing and applications. It has become a hot topic in remote sensing research, drawing significant academic and practical interest [6], [7].

HSI classification mainly consists of feature extraction and classifier classification. Researchers mainly focused on spectral features in the early time. They usually mapped the original spectral features to a new space through linear or nonlinear transformations such as principal component analysis (PCA) [8], linear discriminant analysis (LDA) [9], and manifold learning methods [10], [11]. These methods only used spectral features without considering spatial information, which limited the classification performance. Therefore, spectral-spatial feature extraction techniques have attracted lots of attention, including extended morphological profiles (EMP) [12], extended multi-attribute profiles (EMAP) [13], Gabor filtering [14], sparse representations [15], etc. The commonly used classifiers included support vector machines [16], logistic regression [17], etc. Different combinations of feature extraction techniques and classifiers form a variety of methods.

However, these methods relied on manually crafted features, requiring careful design tailored to specific scenarios. Additionally, their ability to extract high-level image features is also limited [18].

In recent years, the rapid development of deep learning has greatly propelled the advancement of HSI classification research. Deep learning networks can automatically learn high-level and discriminative features based on the data characteristics and have been widely applied in the field of HSI classification [19], [20]. Typical deep learning models include stacked autoencoder (SAE) [21], deep belief networks (DBN) [22], convolutional neural network (CNN) [23], recurrent neural networks (RNNs) [24], etc. From these many models, convolutional neural network has achieved many successes and gained lots of interests [25]. Various CNN architectures have been proposed for extracting spectral and spatial features, including one-dimensional (1D) CNN [26], 2D CNN [27], 1D-2D CNN [28], and 3D CNN [29]. For example, Zhong *et al.* proposed an end-to-end spectral-spatial 3D CNN, deepening the network using residual connections [30]. Zhang *et al.* proposed a 3D dense connected CNN for classification, utilizing dense connections to ensure effective information transmission between layers [31]. Gong *et al.* introduced a novel CNN, which could extract deep multi-scale features from HSI and enhance classification performance by diversified metric [32]. In [33], a double-branch dual-attention mechanism was combined with CNN for enhanced feature extraction. [34] introduced a two-stream residual separable convolution network, specifically designed to address issues of redundant information, data scarcity, and class imbalance typically encountered in HSIs. Researchers have also extensively investigated the integration of CNN with various machine learning techniques, such as transfer learning, ensemble learning, and few-shot learning, to enhance the performance of CNN models under different conditions. Yang *et al.* proposed an effective transfer learning approach to deal with images with different sensors and different number of bands [35]. The proposed method could use multi-sensor HSIs to jointly train a robust and general CNN model. In [36], a pixel-pair feature generation mechanism was specifically designed to augment the training dataset size, while ensemble learning strategies were also employed to mitigate overfitting problem and bolster the robustness of CNN-based classifiers. Yu *et al.* [37] utilized the scheme of prototype learning to address few-shot HSI classification, emphasizing the efficiency in using training samples.

Compared with CNNs which mainly focus on modeling locality, Transformers have demonstrated proficiency in capturing long-range dependencies, enabling a comprehensive understanding of spatial and spectral features' relationships in HSI [38]. Consequently, Transformer-based HSI classification methods have emerged as a promising approach [39], [40]. For instance, He *et al.* proposed HSI-BERT, where each pixel within a given HSI cube sample is treated as a token for the Transformer to capture global context. It was viewed as the initial application of a Transformer-based model for classification with competitive accuracies. [41] Recognizing the significance of long-range dependencies in spectral bands, Hong *et al.* introduced SpectralFormer, utilizing a pure Transformer to process spectral features [42]. Tang *et al.* devised a double-attention Transformer encoder to separately capture spatial and spectral features of HSIs, which were subsequently fused to enhance discriminative capabilities. [43]. In [44], a cross spatial–spectral dense transformer was proposed to extract spatial and spectral features in a dense learning manner. And the cross-attention mechanism was used for feature fusion. In addition to utilizing purely Transformer as described above, researchers have explored the fusion of Transformers with CNNs for feature extraction to leverage the strengths of both models effectively [45]. For example, Sun *et al.* [46] proposed a network comprising 2D and 3D convolution layers to preprocess input HSI samples, followed by a Gaussian-weighted feature tokenizer to generate input tokens for Transformer blocks. To extract multiscale spectral-spatial features in HSIs, Wu *et al.* introduced an enhanced Transformer utilizing multiscale CNNs to generate spectral-spatial tokens with hash-based positional embeddings [47]. Experimental results demonstrated that leveraging multiscale CNN features is advantageous in improving the classification performance. In [48], a lightweight CNN and Transformers in a dual-branch manner are integrated to a basic spectral-spatial block to extract hierarchical features. The proposed model outperformed comparison methods by a significant margin.

However, the traditional Transformer architecture introduces its own set of challenges, primarily due to its quadratic computational complexity driven by the self-attention mechanism [49]. This complexity becomes prohibitive when dealing with the high-dimensional data typical of HSI, which contains both spatial and spectral information. Therefore, fully modeling the long range dependency of spatial-spectral features make a Transform model computationally heavy and, hence, impractical even when compared with other deep learning models. However, structured state space models (SSMs) have been developed to address Transformers' computational inefficiency on long sequences, recently [50]. As an advanced SSM, Mamba has emerged as a promising model which outperforms in both computational efficiency and feature extraction capability [51]. Like Transformers, Mamba models can capture long-range dependencies but with greater computational efficiency, making them well-suited for high-dimensional datasets like HSI.

Therefore, in this paper, we explore the application of the Mamba model for HSI classification and design a spatial-spectral learning framework based on the Mamba model, named spectral-spatial mamba (SS-Mamba). The SS-Mamba model mainly consists of spectral-spatial token generation module and several stacked spectral-spatial Mamba blocks to extract the deep and discriminant features of HSI. The token generation module transforms HSI cubes into sequences of spatial and spectral tokens. These tokens are then processed through several stacked spectral-spatial Mamba blocks. Each spectral-spatial Mamba block employs a double-branch structure that includes spatial mamba feature extraction, spectral mamba feature extraction and spectral-spatial feature enhancement module. The obtained spatial and spectral tokens are processed by the two basic Mamba blocks, respectively. After that, the spectral tokens and spatial tokens cooperates with each other in the designed spectral-spatial feature enhancement module. The dual-branch architecture and spectral-spatial feature enhancement effectively maximize spectral–spatial information fusion and thus improve the performance of HSI classification.

The main contributions are summarized as follows:
1) A spectral-spatial Mamba-based learning framework is proposed for HSI classification, which can effectively utilize Mamba's computational efficiency and powerful long-range feature extraction capability.
2) We designed a spectral-spatial token generation mechanism to convert any given HSI cube to spatial and spectral tokens as sequences for input. It improves and combines the spectral and spatial patch partition to fully exploit the spectral-spatial information contained in HSI samples.
3) A feature enhancement module is designed to enhance the spectral-spatial features and achieve information fusion. By modulating the spatial and spectral tokens using the HSI sample's center region information, the model can focus on the informative region and conduct spectral-spatial information interaction and fusion within each block.

The rest of the paper is organized as follows. Section 2 presents the proposed SS-Mamba for HSI classification. In Section 3, the experiments are conducted on four widely used HSI datasets. The results are presented and discussed. Section 4 gives a discussion about the proposed method. In Section 5, the conclusion is briefly summarized.

**2. Methodology**

The flowchart of the proposed spectral-spatial Mamba model for HSI classification is depicted in Figure 1. It can be seen that the proposed SS-Mamba mainly consists of spectral-spatial token generation module and several stacked spectral-spatial Mamba blocks to extract the deep and discriminant features of HSI. Each spectral-spatial Mamba block employs a double-branch structure that includes spatial mamba feature extraction, spectral mamba feature extraction and spectral-spatial feature enhancement module. To start with, an HSI cube is generated as model's input for any given pixel using spatial neighborhood region, following most of the HSI classification methods. The HSI input samples are then processed to generated spectral and spatial tokens. And these tokens are sent to several spectral-spatial mamba blocks. Each of the blocks is composed of two basic mamba blocks and a spectral-spatial feature enhancement module. The obtained spatial and spectral tokens are processed by the two basic Mamba blocks, respectively. After that, the spectral tokens and spatial tokens cooperates with each other in the designed spectral-spatial feature enhancement module. After feature extraction by these spectral-spatial mamba blocks, the spectral and spatial tokens are averaged and then added to form the final spectral-spatial feature for the given HSI sample. Finally, the obtained feature is sent to a fully connected layer to accomplish classification.

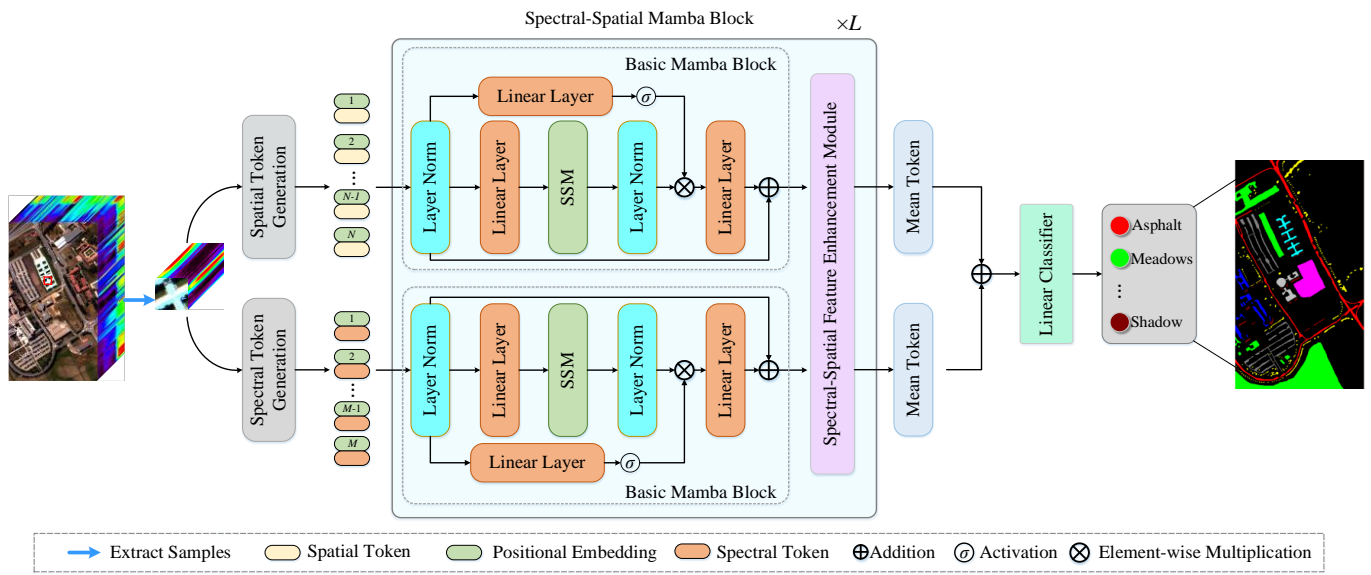

**Figure 1.** The illustration for the design details of the proposed SS-Mamba – working flow. For illustrative purpose, a single image flow instead of a batch is showed here. The proposed SS-Mamba mainly consists of spectral-spatial token generation module and several stacked spectral-spatial Mamba blocks to extract the deep and discriminant features of HSI. And the spectral-spatial information interaction and fusion occurs in the token generation process (early stage), spectral-spatial Mamba blocks (middle stage) and mean token addition (last stage).

*2.1. Overview of the State Space Models*

The state space model serves as a framework for modeling relationship between input and output sequences. Specifically, it maps a one-dimensional input $x(t) \in \mathbb{R}$ to an output $x(t) \in \mathbb{R}$ through the hidden state $h(t) \in \mathbb{R}^N$. This procedure can be formulated through the following ordinary differential equation:

$$h'(t) = \boldsymbol{A}h(t) + \boldsymbol{B}x(t), \tag{1}$$

$$y(t) = \boldsymbol{C}h(t), \tag{2}$$

Here, $\boldsymbol{A} \in \mathbb{R}^{N \times N}$ represents the state matrix, while $\boldsymbol{B} \in \mathbb{R}^{N \times 1}$ and $\boldsymbol{C} \in \mathbb{R}^{N \times 1}$ are the system parameters. To adapt this continuous system for deep learning applications in discrete sequences like image and text, structural state space model (S4) further employs discretization. Specifically, a timescale parameter ∆ is introduced, and a consistent discretization rule is applied to convert **A** and **B** into discrete parameters $\bar{\mathbf{A}}$ and $\bar{\mathbf{B}}$, respectively. The discretization employed in S4 model uses the Zero Order Holding (ZOH) rule, defined as:

$$\bar{\mathbf{A}} = e^{\Delta \mathbf{A}}, \tag{3}$$

$$\bar{\mathbf{B}} = (\Delta \mathbf{A})^{-1}(e^{\Delta \mathbf{A}} - \mathbf{I}) - \Delta \mathbf{B}, \tag{4}$$

And the discretization SSM can then be calculated in recurrence denoted by the following equations:

$$h_t = \bar{\mathbf{A}} h_{t-1} + \bar{\mathbf{B}} x_t, \tag{5}$$

$$y_t = \mathbf{C} h_t. \tag{6}$$

For fast and efficient parallel training, the above process in recurrence can also be reformulated using convolution:

$$\mathbf{y} = \mathbf{x} * \bar{\mathbf{K}}, \tag{7}$$

where $\bar{\mathbf{K}}$ denotes the structured convolutional kernels. It is obtained by

$$\bar{\mathbf{K}} = (\mathbf{C}\bar{\mathbf{B}}, \mathbf{C}\bar{\mathbf{A}}\bar{\mathbf{B}}, \dots, \mathbf{C}\bar{\mathbf{A}}^{N-1}\bar{\mathbf{B}}), \tag{8}$$

where $N$ denoted the input sequence's length.

In Mamba, the selective scan mechanism is further used. Specifically, the matrices **B**, **C** and ∆ are generated from the input data **x**, allowing the model to dynamically modeling the contextual relationship within the input sequence. Inspired by the Transformer and Hungry Hungry Hippo (H3) architectures [52], the normalization, residual connectivity, gated MLP, and SSM are combined to form a basic Mamba block, which constitutes the fundamental component of a Mamba network. The structure of a basic Mamba block is shown in Figure 1.

### 2.2. Spectral-Spatial Tokens Generation

To utilize mamba's power sequence modeling ability, the HSI input is need to be transformed to sequence as Mamba's input. Given an HSI training sample $\mathbf{x} \in \mathbb{R}^{H \times W \times B}$, where $H$ and $W$ are the height and width of the input, respectively, and $B$ is the number of spectral bands, the proposed method converts it to spatial tokens sequence and spatial tokens sequence, respectively. Figure 2 illustrates the spectral and spatial token generation process.

For spatial token generation, there are mainly three steps including spectral mapping, spatial partition and patch embedding.

We firstly process spectral features before performing spatial partition for the spatial Mamba. The purpose is to fully exploit the spectral-spatial information contained in HSI samples in the case that the spatial partition-based mamba feature extraction mainly focuses on spatial tokens' relationships. Specifically, the input HSI sample is firstly reshaped to a tensor with shape of $HW \times B$, and then a lightweight multilayer perception (MLP) is used for spectral feature mapping:

$$\mathbf{x}_{spa} \in \mathbb{R}^{HW \times D'} = \text{MLP}(\text{Reshape}(\mathbf{x})), \tag{9}$$

where MLP(·) denotes the corresponding mapping function and $D'$ is the mapped feature dimension.

The spectrally mapped HSI sample is then spatially partitioned into $N = HW/P_{spa}^2$ non-overlapped patches $\bar{\mathbf{p}}_{spa}$, where $\bar{\mathbf{p}}_{spa} \in \mathbb{R}^{N \times P^2 D'}$ and $P_{spa}$ is the spatially partitioned patch size. After that, a linear layer is used to project these patches to a given dimension and obtained spatial input tokens:

$$\mathbf{Z}_{spa}^0 = \bar{\mathbf{p}}_{spa}\mathbf{E}_{spa}, \quad (10)$$

where $\mathbf{E}_{spa} \in \mathbb{R}^{P^2 D' \times D}$ denotes the learnable matrix in the linear layer and $D$ represents the tokens' dimension.

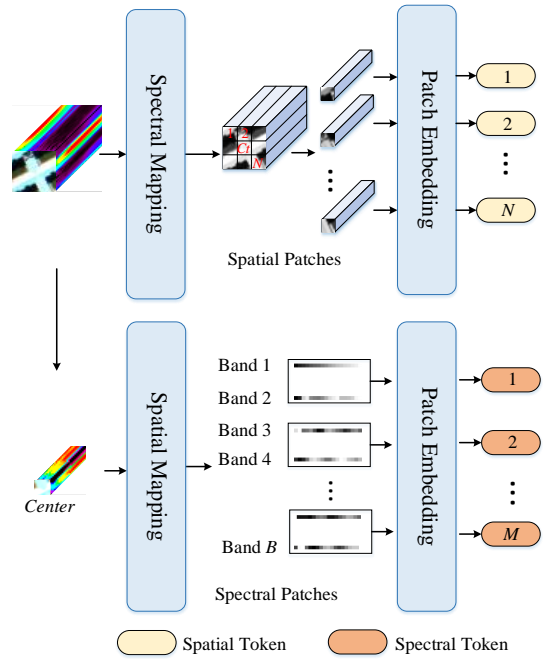

**Figure 2.** The procedure of spectral and spatial tokens generation. The spatial branch processes the entire region by partitioning it along the spatial dimension after spectral mapping, while the spectral branch focuses on the center region, partitioning it along the spectral dimension after spatial mapping.

Similar to spatial token generation, the spectral token generation process also contains three steps including spatial mapping, spectral partition and patch embedding.

To start with, we firstly extract the center region as the input $\hat{\mathbf{x}} \in \mathbb{R}^{S \times S \times B}$, where $S$ is a small integer, i.e., 3. A small center region can make the spectral feature robust when compared with only center pixel. Like the spatial token generation does, spatial feature is processed before performing spatial partition. The obtained input $\hat{\mathbf{x}}$ is reshaped to a tensor with shape of $B \times S^2$, and then another lightweight MLP is used for spatial feature mapping:

$$\mathbf{x}_{spe} \in \mathbb{R}^{B \times D'} = \text{MLP}(\text{Reshape}(\hat{\mathbf{x}})). \quad (11)$$

The spatially mapped HSI sample $\mathbf{x}_{spe}$ is then spectral partitioned into $M = B/P_{spe}$ non-overlapped patches $\bar{\mathbf{p}}_{spe}$, where $\bar{\mathbf{p}}_{spe} \in \mathbb{R}^{M \times P_{spe} D'}$ and $P_{spe}$ is the spectrally partitioned patch size. After that, the spectral input token can be obtained by patch embedding:

$$\mathbf{Z}_{spe}^0 = \bar{\mathbf{p}}_{spe}\mathbf{E}_{spe} \in \mathbb{R}^{M \times D}, \quad (12)$$

where $\mathbf{E}_{spe} \in \mathbb{R}^{P_{spe} D' \times D}$ denotes the learnable matrix in the linear layer.

Following vision Transformer does, the obtained tokens are added with positional embedding to provide location information within the HSI sample. It is worth noting that the spatial tokens are provided with 2D sinusoidal position embedding, while the spectral tokens using 1D positional embedding.

### 2.3. Spectral-Spatial Mamba Block

The spectral-spatial feature extraction is mainly achieved by several spectral-spatial mamba blocks, which consists of two distinct basic mamba blocks and a feature enhancement module. The two basic mamba blocks are mainly used for spatial and spectral feature extraction, respectively. And in the feature enhancement module, the spatial or spectral tokens are modulated using the HSI sample's center region information from the tokens of the other type. By this way, the spectral tokens and spatial tokens cooperates with each other and achieve information fusion in each blocks.

Specifically, in the $l$-th block, the spectral and spatial tokens are firstly processed by mamba blocks:

$$\mathbf{Z}_{spe}^l = MB_{spe}^l(\mathbf{Z}_{spe}^{l-1}), \tag{13}$$

$$\mathbf{Z}_{spa}^l = MB_{spa}^l(\mathbf{Z}_{spa}^{l-1}), \tag{14}$$

where $\mathbf{Z}_{spe}^l$ and $\mathbf{Z}_{spa}^l$ denote the $l$-th block's output spectral and spatial tokens, respectively. And $MB_{spe}^l$ and $MB_{spa}^l$ represents the mapping function of basic mamba blocks introduced in Section II. A for spectral and spatial tokens, respectively.

After the mamba's feature extraction, the obtained spectral and spatial tokens are then sent to the feature enhancement module for information interaction, just as Figure 3 shows.

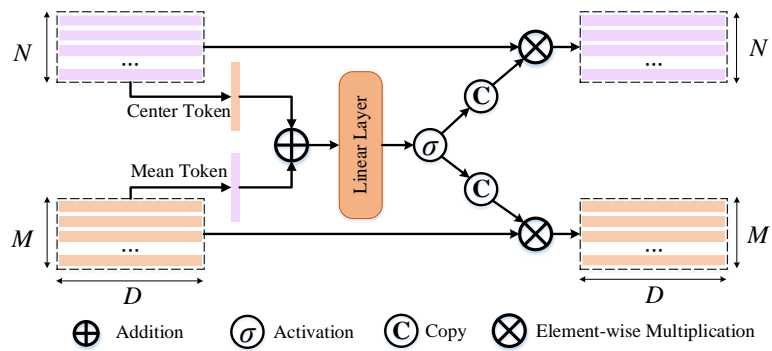

**Figure 3.** The illustration of spectral-spatial feature enhancement module.

For the spatial tokens, we first take out the token corresponding to the central patch, denoting as $\mathbf{f}_1^l$:

$$\mathbf{f}_1^l = \mathbf{Z}_{spa}^l[(N+1)/2, :], \tag{15}$$

And for the spectral tokens, these $N$ tokens are averaged to obtain the center region's spectral feature from the view of the spectral branch, denoted as $\mathbf{f}_2^l$:

$$\mathbf{f}_2^l = \frac{1}{M}\sum_{j=1}^M \mathbf{Z}_{spe}^l[j, :] \qquad j = 1, \cdots, M. \tag{16}$$

The obtained features are then fused by average:

$$\mathbf{f}^l = (\mathbf{f}_1^l + \mathbf{f}_2^l)/2. \tag{17}$$

An MLP is used for further feature extraction and the *sigmoid* activation is also used for scaling:

$$\mathbf{s}^l = \sigma(\text{MLP}(\mathbf{f}^l)), \tag{18}$$

where $\sigma$ represents the *sigmoid* activation function. The obtained feature $\mathbf{s}^l$ can be viewed as the modulated weights, which is going to be multiplied with the spectral and spatial tokens. Before multiplication, it should cope with matching dimension. We make $N$ and $M$ copies of $\mathbf{s}$ for spatial and spatial tokens, respectively:

$$\mathbf{A}_{spa}^{l} = [\mathbf{s}^l; \mathbf{s}^l; \cdots \mathbf{s}^l] \in \mathbb{R}^{N \times D}, \tag{19}$$

$$\mathbf{A}_{spe}^{l} = [\mathbf{s}^l; \mathbf{s}^l; \cdots \mathbf{s}^l] \in \mathbb{R}^{M \times D}. \tag{20}$$

The obtained modulated matrices $\mathbf{A}_{spa}^{l}$ and $\mathbf{A}_{spe}^{l}$ are then multiplied with the spectral and spatial tokens for forcing the model to focus more on the most informative central region as the input HSI image's labels is mainly determined by the center pixel's label and the spectral information is mainly needed here for enhancement. The procedure can be formulated as:

$$\mathbf{Z}_{spa}^{l} = \mathbf{A}_{spa}^{l} \odot \mathbf{Z}_{spa}^{l}, \tag{21}$$

$$\mathbf{Z}_{spe}^{l} = \mathbf{A}_{spe}^{l} \odot \mathbf{Z}_{spe}^{l}, \tag{22}$$

where $\odot$ means the element-wise multiplication.

After feature extraction by $L$ spectral-spatial mamba blocks, the obtained spectral and spatial tokens are averaged and then added to form the final spectral-spatial feature for the given HSI sample, which can be formulated by:

$$\mathbf{f}_{spa} = \frac{1}{N} \sum_{j=1}^{M} \mathbf{Z}_{spa}^{L}[j,:] \qquad j = 1, \cdots, N, \tag{23}$$

$$\mathbf{f}_{spe} = \frac{1}{M} \sum_{j=1}^{M} \mathbf{Z}_{spe}^{L}[j,:] \qquad j = 1, \cdots, M, \tag{24}$$

$$\mathbf{f} = \mathbf{f}_{spa} + \mathbf{f}_{spe}. \tag{25}$$

And the representation $\mathbf{f}$ of the input HSI sample is sent into a fully connected layer to obtain the final logit prediction:

$$\hat{\mathbf{y}} \in \mathbb{R}^K = \text{FC}(\mathbf{f}), \tag{26}$$

where $\text{FC}(\cdot)$ denotes the mapping function of a fully connected layer and $\hat{\mathbf{y}}$ is the predicted label vector for the input HSI sample $\mathbf{x}$. Besides, $K$ is the number of classes.

The cross-entropy loss is used for optimizing the designed model:

$$\mathcal{L}_{cls} = -\sum_{k=1}^{K} y_k \log \hat{y}_k, \tag{27}$$

where $\hat{y}_k$ is the $k$-th element of $\hat{\mathbf{y}}$, and $y_k$ is the $k$-th element of the one-hot label vector.

## 3. Results

### 3.1. Datasets Description

The performance of the proposed mamba-based method is assessed using four widely-used HSI datasets: Indian Pines, Pavia University, Houston, and Chikusei.

1) Indian Pines: The Indian Pines dataset primarily records agricultural areas in Northwestern Indiana, USA, captured in June 1992 by the Airborne Visible/Infrared Imaging Spectrometer (AVIRIS). The dataset comprises 145×145 pixels with a spatial resolution of 20 meters. It includes 220 spectral bands, covering wavelengths from 400 nm to 2500 nm. For application, 200 bands are retained after removing 20 bands with low signal-to-noise ratio. The dataset includes 10,249 labeled samples belonging to 16 distinct land-cover categories. The dataset's false color and ground-truth maps are illustrated in Figure 4, and the distribution of training and test samples is detailed in Table 1.

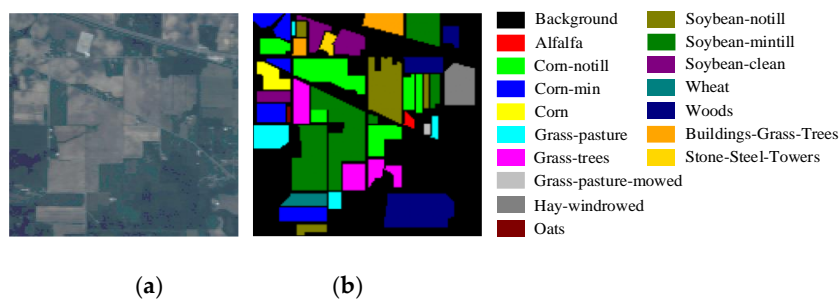

(**a**) (**b**)

**Figure 4.** Indian Pines dataset: (**a**) true color map (645nm, 547nm, 458nm), (**b**) ground truth.

**Table 1.** Land cover classes and numbers of samples in the Indian Pines dataset.

| No. | Class Name | Training samples | Test samples | Total samples |
|-----|------------|------------------|--------------|---------------|
| 1 | Alfalfa | 20 | 26 | 46 |
| 2 | Corn-notill | 20 | 1408 | 1428 |
| 3 | Corn-mintill | 20 | 810 | 830 |
| 4 | Corn | 20 | 217 | 237 |
| 5 | Grass-pasture | 20 | 463 | 483 |
| 6 | Grass-trees | 20 | 710 | 730 |
| 7 | Grass-pasture-mowed | 20 | 8 | 28 |
| 8 | Hay-windrowed | 20 | 458 | 478 |
| 9 | Oats | 15 | 5 | 20 |
| 10 | Soybean-notill | 20 | 952 | 972 |
| 11 | Soybean-mintill | 20 | 2435 | 2455 |
| 12 | Soybean-clean | 20 | 573 | 593 |
| 13 | Wheat | 20 | 185 | 205 |
| 14 | Woods | 20 | 1245 | 1265 |
| 15 | Buildings-Grass-Trees | 20 | 366 | 386 |
| 16 | Stone-Steel-Towers | 20 | 73 | 93 |
| | Total | 315 | 9934 | 10249 |

2) Pavia University: The Pavia University dataset, captured by the Reflective Optics System Imaging Spectrometer (ROSIS) over the University of Pavia, Italy, contains 610 × 340 pixels at a fine spatial resolution of 1.3 m. This dataset records 103 spectral bands and labels 42,776 pixels from nine land-cover classes. Figure 5 shows the false color and ground-truth maps, while Table 2 shows the distribution of training and test samples of each class.

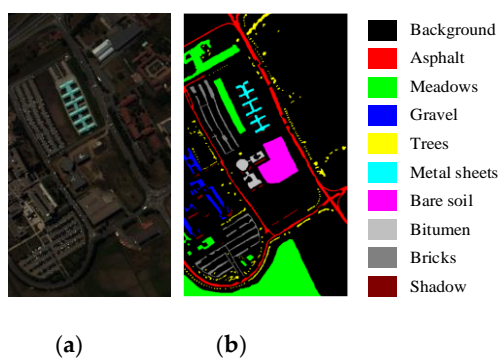

(**a**) (**b**)

**Figure 5.** Pavia University dataset: (**a**) false color map (643nm, 547nm, 455nm), (**b**) ground truth.

Table 2. Land cover classes and numbers of samples in the Pavia University dataset.

| No. | Class Name | Training samples | Test samples | Total samples |
|---|---|---|---|---|
| 1 | Asphalt | 20 | 6611 | 6631 |
| 2 | Meadows | 20 | 18629 | 18649 |
| 3 | Gravel | 20 | 2079 | 2099 |
| 4 | Trees | 20 | 3044 | 3064 |
| 5 | Mental sheets | 20 | 1325 | 1345 |
| 6 | Bare soil | 20 | 5009 | 5029 |
| 7 | Bitumen | 20 | 1310 | 1330 |
| 8 | Bricks | 20 | 3662 | 3682 |
| 9 | Shadow | 20 | 927 | 947 |
|  | Total | 180 | 42596 | 42776 |

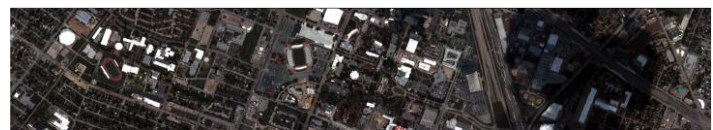

(a)

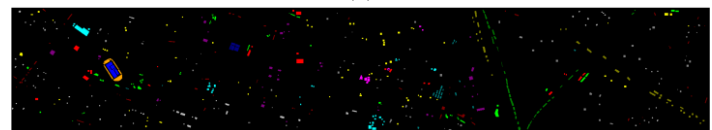

(b)

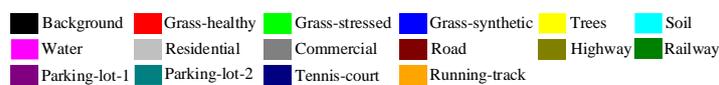

**Figure 6.** Houston dataset: (**a**) true color map (650nm, 550nm, 450nm), (**b**) ground truth.

Table 3. Land cover classes and numbers of samples in the Houston dataset.

| No. | Class Name | Training samples | Test samples | Total samples |
|---|---|---|---|---|
| 1 | Grass-healthy | 20 | 1231 | 1251 |
| 2 | Grass-stressed | 20 | 1234 | 1254 |
| 3 | Grass-synthetic | 20 | 677 | 697 |
| 4 | Tree | 20 | 1224 | 1244 |
| 5 | Soil | 20 | 1222 | 1242 |
| 6 | Water | 20 | 305 | 325 |
| 7 | Residential | 20 | 1248 | 1268 |
| 8 | Commercial | 20 | 1224 | 1244 |
| 9 | Road | 20 | 1232 | 1252 |
| 10 | Highway | 20 | 1207 | 1227 |
| 11 | Railway | 20 | 1215 | 1235 |
| 12 | Parking-lot-1 | 20 | 1213 | 1233 |
| 13 | Parking-lot-2 | 20 | 449 | 469 |
| 14 | Tennis-court | 20 | 4008 | 428 |
| 15 | Running-track | 20 | 640 | 660 |
|  | Total | 300 | 14729 | 15029 |

3) Houston: The Houston dataset focuses on an urban area around the University of Houston campus, USA. It was captured by the National Center for Airborne Laser Mapping (NCALM) for the 2013 IEEE GRSS Data Fusion Contest [53]. The dataset contains 349 × 1905 pixels with a spatial resolution of 2.5 m per pixel and includes 144 spectral bands spanning 380 nm to 1050 nm. It consists of 15,029 labeled pixels categorized into 15 land-cover classes. Figure 6 illustrates both the false color and ground-truth maps, and Table 3 details the class-wise distribution of training and test samples.

4) Chikusei: The Chikusei dataset is an aerial hyperspectral dataset acquired using the Headwall Hyperspec-VNIR-C sensor in Chikusei, Japan, on July 29, 2014 [54]. It comprises 128 spectral bands spanning wavelengths from 343 nm to 1018 nm. The dataset spatial size is 2517×2335, with a resolution of 2.5 meters. It includes 19 types of land-covers, encompassing urban and rural areas. In Figure 7, the true color composite image and the corresponding ground truth map of the Chikusei dataset are depicted. Table 4 shows the distribution of training and test samples of each class.

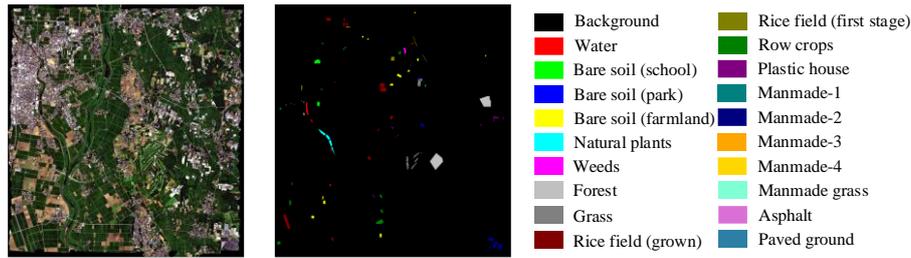

**Figure 7.** Chikusei dataset: (**a**) false color map (648nm, 549nm, 448nm), (**b**) ground truth.

**Table 4.** Land cover classes and numbers of samples in the Chikusei dataset.

| No. | Class Name | Training samples | Test samples | Total samples |
|---|---|---|---|---|
| 1 | Water | 5 | 2840 | 2845 |
| 2 | Bare soil (school) | 5 | 2854 | 2859 |
| 3 | Bare soil (park) | 5 | 281 | 286 |
| 4 | Bare soil (farmland) | 5 | 4847 | 4852 |
| 5 | Natural plants | 5 | 4292 | 4297 |
| 6 | Weeds | 5 | 1103 | 1108 |
| 7 | Forest | 5 | 20511 | 20516 |
| 8 | Grass | 5 | 6510 | 6515 |
| 9 | Rice field (grown) | 5 | 13364 | 13369 |
| 10 | Rice field (first stage) | 5 | 1263 | 1268 |
| 11 | Row crops | 5 | 5956 | 5961 |
| 12 | Plastic house | 5 | 2188 | 2193 |
| 13 | Manmade-1 | 5 | 1215 | 1220 |
| 14 | Manmade-2 | 5 | 7659 | 7664 |
| 15 | Manmade-3 | 5 | 426 | 431 |
| 16 | Manmade-4 | 5 | 217 | 222 |
| 17 | Manmade grass | 5 | 1035 | 1040 |
| 18 | Asphalt | 5 | 796 | 801 |
| 19 | Paved ground | 5 | 140 | 145 |
| | Total | 95 | 77497 | 77592 |

*3.2. Experimental Setup*

To demonstrate the effectiveness of the proposed SS-Mamba, kinds of HSI classification methods are selected and evaluated as comparison methods. These comparison methods are listed as follows.

1) EMP-SVM [16]: this method utilizes EMP for spatial feature extraction followed by a classic SVM for final classification. This approach is commonly employed as a benchmark against deep learning-based methodologies.

2) CNN: it is a vanilla CNN which simply contains four convolutional layers. It is viewed as a basic spectral-spatial deep learning-based model for HSI classification.
3) SSRN [30]: it is a 3D deep learning framework which uses three-dimensional convolutional kernels and residual blocks to improve the CNN's performance of HSI classification.
4) DBDA [33]: it is an advanced CNN model that integrates a double-branch dual-attention mechanism for enhanced feature extraction. It is used for comparison with Transformer-based methodologies, which rely on self-attention mechanisms.
5) MSSG [54]: it employs a super-pixel structured graph U-Net to learn multiscale features across multilevel graphs. As a graph CNN and global learning model, MSSG is contrasted with the proposed Mamba and patch-based methods.
6) SSFTT [46]: SSFTT is a spatial-spectral Transformer that designs a unique tokenization method and use CNN to provide local features for Transformer.
7) LSFAT [55]: it is a local semantic feature aggregation-based transformer which has the advantages of learning multiscale features.
8) CT-Mixer [45]: it is an aggregated framework of CNN and Transformer, which is hoped to utilized both the advantages of the above two classic models effectively.

The listed comparison methods encompass a diverse kind of methods, including traditional method, CNN-based methods, Transformer-based models, attention-based methods, and mixture methods, and thus offers a comprehensive and exhaustive basis for comparison. For details, one can refer to the original papers.

When using SS-Mamba, a spatial window size of 27×27 (i.e., $H = W = 27$) is used for any given pixel to generate the model's HSI input sample. As for the hyper-parameters in the proposed SS-Mamba, the spatial partition size is set to three, that is $P_{spa}$ =3. And the spectral partition size is set to two, that is $P_{spe}$ =2. The embedding dimension for each mamba block is 64, that is $D$ =64. The widely-used multi-step learning rate scheduler is employed for training. Specifically, the initial learning rate is 0.0005, and it is divided by two every 80 epochs. The total number of epochs is set to 180 for all the four datasets. And the mini-batch strategy with a batch size of 256 is used for all the datasets.

The classification performance is primarily evaluated based on metrics including overall accuracy (OA), average accuracy (AA), and the Kappa coefficient (K). To ensure the reliability and robustness of the classification methods' output results, the experiments are repeated ten times with varying random initializations. Besides, the training samples are randomly selected from all labeled samples each time. However, the training samples remain the same for all the involved methods each time.

*3.3. Ablation Experiments*

*3.3.1 Ablation Experiment with Basic Sequence Model*

To demonstrate the classification ability of Mamba model, the ablation experiments with different sequence models are conducted on the four datasets. These sequence models include long short-term memory network (LSTM), gated recurrent unit (GRU), and Transformer. The overall framework of the model remains unchanged, with only the basic sequence model varying. The related results are showed in Tables 5-8. From the obtained results, one can see that

i) Spectral-spatial models achieve the highest accuracies with the same basic sequence model, followed by the spatial model, with the spectral model proved to be the least effective. For example, in Table 5, spectral-spatial LSTM achieves better results than spatial LSTM and spectral LSTM, with improvements of 3.76 percentage points and 32.85 percentage points in terms of OA, respectively. Besides, the spectral-spatial mamba outperforms spatial mamba by 2.77 percentage points, 5.14 percentage points, and 0.0364 in terms of OA, AA, and K on the Pavia University dataset, respectively. The results indicate that the designed spectral-spatial learning framework is effective for different sequence models.

Table 5. Ablation experiment with different sequence model on the Indian Pines dataset

|  |  | LSTM | GRU | Transformer | Mamba |
|---|---|---|---|---|---|
| Spectral Only | OA (%) | 56.93 | 59.41 | 54.53 | **60.95** |
|  | AA (%) | 69.84 | 72.09 | 67.50 | **73.31** |
|  | K×100 | 51.81 | 54.65 | 49.12 | **56.27** |
| Spatial Only | OA (%) | 86.02 | 84.39 | 83.60 | **87.40** |
|  | AA (%) | 91.17 | 90.17 | 89.41 | **92.35** |
|  | K×100 | 84.15 | 82.31 | 81.43 | **85.71** |
| Spectral-Spatial | OA (%) | 89.78. | 90.38 | 88.03 | **91.59** |
|  | AA (%) | 94.15 | 94.74 | 93.18 | **95.46** |
|  | K×100 | 88.24 | 89.05 | 86.38 | **90.42** |

Table 6. Ablation experiment with different sequence model on the Pavia University dataset

|  |  | LSTM | GRU | Transformer | Mamba |
|---|---|---|---|---|---|
| Spectral Only | OA (%) | 73.40 | 73.53 | 74.55 | **75.52** |
|  | AA (%) | 81.51 | 81.83 | 81.95 | **83.11** |
|  | K×100 | 66.34 | 66.55 | 67.78 | **68.88** |
| Spatial Only | OA (%) | 89.39 | 90.65 | 92.62 | **93.63** |
|  | AA (%) | 89.11 | 90.05 | 95.28 | **93.49** |
|  | K×100 | 86.24 | 87.84 | 90.36 | **91.67** |
| Spectral-Spatial | OA (%) | 95.51 | 95.95 | 94.99 | **96.40** |
|  | AA (%) | 97.97 | 97.91 | 97.26 | **98.43** |
|  | K×100 | 94.17 | 94.85 | 93.47 | **95.31** |

Table 7. Ablation experiment with different sequence model on the Houston dataset

|  |  | LSTM | GRU | Transformer | Mamba |
|---|---|---|---|---|---|
| Spectral Only | OA (%) | 81.32 | 83.81 | 84.61 | **84.86** |
|  | AA (%) | 82.64 | 85.17 | 85.70 | **85.92** |
|  | K×100 | 79.81 | 82.51 | 83.37 | **83.51** |
| Spatial Only | OA (%) | 87.88 | 88.52 | 89.16 | **90.21** |
|  | AA (%) | 89.42 | 89.87 | 90.20 | **91.31** |
|  | K×100 | 86.91 | 87.60 | 88.28 | **89.42** |
| Spectral-Spatial | OA (%) | 93.50 | 93.81 | 93.38 | **94.30** |
|  | AA (%) | 94.17 | 94.32 | 93.48 | **94.96** |
|  | K×100 | 92.97 | 93.24 | 92.84 | **93.84** |

Table 8. Ablation experiment with different sequence model on the Chikusei dataset

|  |  | LSTM | GRU | Transformer | Mamba |
|---|---|---|---|---|---|
| Spectral Only | OA (%) | 68.73 | 70.00 | 68.28 | **78.38** |
|  | AA (%) | 79.80 | 81.73 | 82.68 | **85.58** |
|  | K×100 | 64.38 | 65.83 | 64.27 | **75.39** |
| Spatial Only | OA (%) | 92.01 | 93.30 | 93.13 | **93.83** |
|  | AA (%) | 92.76 | 93.77 | 93.37 | **93.84** |
|  | K×100 | 90.85 | 92.30 | 92.13 | **92.92** |
| Spectral-Spatial | OA (%) | 94.31 | 94.38 | 94.21 | **94.97** |
|  | AA (%) | 94.18 | 94.21 | 94.02 | **94.83** |
|  | K×100 | 93.59 | 93.62 | 93.54 | **94.22** |

ii) With the learning framework, the Mamba-based models achieve higher accuracies than the classical sequence models such as LSTM, GRU, and Transformer. For examples, the spectral-spatial mamba outperforms spectral-spatial GRU by 0.45 percentage points, 0.52 percentage points, and 0.0046 in terms of OA, AA, and K on the Pavia University dataset, respectively. On the Houston dataset, spectral-spatial mamba yields better results than Transformer, GRU, and LSTM, with improvements of 0.92 percentage points, 0.49 percentage points, and 0.80 percentage points for OA, respectively. one can also draw similar conclusion for spatial or spectral learning methods on the four datasets. The results indicate that the used Mamba-based sequence models are effective for different learning frameworks.

Notably, we find that all the spectral models seem harder to be trained well compared with the other models, especially on the Indian Pines dataset. Specifically, the spectral models need larger learning rates and more epochs to train. Besides, these models are all much harder to train and perform worst on the Indian Pines dataset, probably due to the low quality of the dataset.

*3.3.2 Ablation Experiment with Feature Enhancement Module*

For spectral-spatial Mamba model, we have designed the feature enhancement module, which can use the HSI sample's center region information from the tokens to enhance spectral-spatial features. To demonstrate the effectiveness of this module, the ablation experiment is conducted. And Table 9 shows the ablation experimental results. From the results, one can see that the spectral-spatial Mamba with feature enhancement module achieves higher accuracies when compared with the model without feature enhancement module. For example, the feature enhancement module improves the classification results by 2.09 percentage points, 1.78 percentage points, and 0.0226 in terms of OA, AA, and K on the Houston dataset, respectively. The results indicate that the designed feature enhancement module is effective.

**Table 9.** Ablation experiment for the feature enhancement module

|  | Indian Pines | | Pavia University | | Houston | |
| --- | --- | --- | --- | --- | --- | --- |
|  | w/ | w/o | w/ | w/o | w/ | w/o |
| OA (%) | **91.59** | 89.01 | **96.40** | 95.97 | **94.30** | 92.21 |
| AA (%) | **95.46** | 93.35 | **98.43** | 98.04 | **94.96** | 93.18 |
| K×100 | **90.42** | 87.51 | **95.31** | 94.75 | **93.84** | 91.58 |

w/: with, w/o: without

*3.4. Classification results*

The classification results of different methods on the four datasets are shown in Tables 10-13. From these results, one can see that the proposed SS-Mamba have achieved superior classification performance over other comparison methods when using twenty training samples per class across all the four datasets.

Specifically, it can be seen that 1) The CNN-based methods usually achieve higher classification accuracies than the Transformer-based methods. As an example, on the Pavia University dataset, the MSSG performs better than the CT-Mixer with an improvement of 1.46 percentage points for OA, 3.44 percentage points for AA, and 0.0194 for K, respectively. On the Houston dataset, the overall accuracies of the comparison Transformer-based methods are lower that 93%. In contrast, the overall accuracies achieved by CNN-based methods are generally higher than 93%. And the OA of MSSG is even close to 94%. It seems that it is necessary to improve Transformer-based methods in the case of limited training sample. 2) The designed spectral-spatial Mamba model obtains highest accuracies when compared with other methods on the four datasets. For example, in Table 9, SS-Mamba yields better results than the DBDA, with an improvement of 0.53 percentage points for OA, 3.68 percentage points for AA, and 0.0073 for K. On the Houston dataset, the proposed method achieves better results than MSSG, DBDA, and SSFTT, with improvements of 0.38 percentage points, 0.63 percentage points, and 4.12 percentage points in terms of OA, respectively. Compared with Transformer, Mamba, which is also a sequence model, has better classification performance, which shows the potential application of sequence model in HSI classification.

**Table 10.** Testing data classification results (mean ± standard deviation) on the Indian Pines dataset

| Class | EMP-SVM | CNN | SSRN | DBDA | MSSG | LSFAT | SSFTT | CT-Mixer | **SS-Mamba** |
|---|---|---|---|---|---|---|---|---|---|
| Alfalfa | 96.54±2.07 | **100.0±0.00** | 86.04±9.73 | 79.35±12.80 | 98.46±2.55 | **100.0±0.00** | **100.0±0.00** | **100.0±0.00** | **100.0±0.00** |
| Corn-notill | 63.74±7.45 | 78.96±6.22 | 86.85±6.00 | 86.89±10.74 | **87.42±6.17** | 81.13±6.11 | 85.38±3.92 | 81.94±3.95 | 80.30±4.58 |
| Corn-mintill | 76.56±4.40 | 87.75±5.96 | 86.54±5.65 | 86.88±8.79 | 87.80±6.91 | 86.90±5.32 | 87.63±5.69 | 87.47±4.29 | **88.54±5.07** |
| Corn | 81.94±5.14 | 98.53±2.56 | 73.29±13.10 | 81.64±11.47 | 96.22±7.07 | 99.07±1.23 | 98.06±2.43 | 99.35±1.11 | **99.49±0.76** |
| Grass-pasture | 86.57±3.85 | 92.18±3.21 | **98.00±1.96** | 97.02±3.74 | 88.70±6.28 | 92.89±2.90 | 94.15±2.09 | 93.24±1.54 | 93.69±1.99 |
| Grass-trees | 92.85±4.63 | 95.41±2.53 | 97.79±1.40 | 96.54±1.25 | 97.69±1.78 | 95.35±3.89 | 97.98±1.26 | 95.30±3.47 | **98.34±1.53** |
| Grass-pasture-mowed | 92.50±6.12 | **100.0±0.00** | 73.78±18.70 | 58.08±28.22 | **100.0±0.00** | **100.0±0.00** | **100.0±0.00** | **100.0±0.00** | **100.0±0.00** |
| Hay-windrowed | 94.10±3.25 | 99.80±0.59 | 99.46±0.83 | 99.62±0.58 | 99.93±0.20 | 99.56±0.97 | 99.34±1.22 | 99.98±0.07 | **100.0±0.00** |
| Oats | 98.00±6.00 | **100.0±0.00** | 44.13±20.56 | 32.65±11.22 | **100.0±0.00** | **100.0±0.00** | **100.0±0.00** | **100.0±0.00** | **100.0±0.00** |
| Soybean-notill | 68.28±7.20 | 88.93±3.10 | 79.53±6.45 | 81.08±8.08 | 84.79±8.70 | 87.58±4.89 | **89.66±4.47** | 87.88±4.44 | 89.61±3.92 |
| Soybean-mintill | 59.22±4.69 | 87.41±5.47 | 92.94±3.50 | **95.43±4.19** | 86.75±8.83 | 85.26±3.98 | 84.72±5.24 | 86.89±3.17 | 89.51±5.87 |
| Soybean-clean | 66.61±7.68 | 83.84±6.23 | 82.78±10.73 | 90.57±13.45 | 83.58±12.48 | 83.70±4.94 | 77.40±4.99 | 82.57±6.87 | **93.32±4.74** |
| Wheat | 97.08±1.81 | 99.57±0.89 | 95.49±3.82 | 91.42±5.35 | **100.0±0.00** | 99.89±0.32 | 98.86±2.15 | 98.97±1.89 | 99.46±1.62 |
| Woods | 86.39±5.68 | 97.39±1.71 | 97.87±1.14 | 98.43±1.05 | **99.32±0.96** | 97.69±1.18 | 98.75±1.18 | 97.07±1.48 | 98.63±1.40 |
| Buildings-Grass-Trees | 71.48±8.55 | 97.19±2.60 | 87.30±7.75 | 88.99±5.46 | **97.32±3.96** | 95.63±3.74 | 96.91±2.02 | 96.97±2.59 | 96.67±3.16 |
| Stone-Steel-Towers | 95.48±4.71 | 99.18±0.91 | 79.51±4.48 | 68.09±12.75 | 99.32±0.68 | 98.49±1.67 | **100.0±0.00** | 99.59±0.63 | 99.73±0.82 |
| OA (%) | 73.32±2.25 | 89.62±1.72 | 89.44±1.38 | 90.26±3.06 | 90.60±2.03 | 89.35±1.38 | 90.10±1.58 | 89.87±1.28 | **91.59±1.85** |
| AA (%) | 82.96±1.31 | 94.14±0.72 | 85.08±2.08 | 83.29±2.47 | 94.21±1.46 | 93.95±0.62 | 94.30±0.83 | 94.20±0.72 | **95.46±0.90** |
| K×100 | 69.93±2.50 | 88.20±1.93 | 88.00±1.55 | 88.95±3.43 | 89.27±2.30 | 87.90±1.55 | 88.76±1.78 | 88.49±1.44 | **90.42±2.08** |

**Table 11.** Testing data classification results (mean ± standard deviation) on the Pavia University dataset

| Class | EMP-SVM | CNN | SSRN | DBDA | MSSG | LSFAT | SSFTT | CT-Mixer | **SS-Mamba** |
|---|---|---|---|---|---|---|---|---|---|
| Asphalt | 81.27±6.60 | 90.85±4.53 | 97.84±1.71 | **98.74±1.20** | 97.06±3.69 | 87.16±3.78 | 86.16±6.15 | 90.59±5.26 | 95.70±3.41 |
| Meadows | 83.13±3.26 | 93.83±3.49 | 97.72±0.81 | **99.51±0.36** | 92.29±5.91 | 95.05±3.17 | 94.50±3.60 | 94.62±4.88 | 94.05±4.27 |
| Gravel | 81.60±4.51 | 98.12±1.27 | 83.71±8.17 | 90.81±12.10 | **99.97±0.10** | 93.95±4.43 | 93.94±4.13 | 94.39±4.19 | 99.61±0.54 |
| Trees | 95.29±2.44 | 96.29±1.39 | 97.70±2.01 | 92.74±7.92 | 97.12±1.38 | 92.76±4.46 | 88.97±5.25 | 84.61±6.80 | **98.92±0.55** |
| Mental sheets | 99.26±0.26 | 99.33±0.52 | 99.86±0.27 | 99.53±0.63 | **100.0±0.00** | 98.85±0.89 | 98.95±1.06 | 99.18±0.67 | **100.0±0.00** |
| Bare soil | 80.27±6.31 | 99.47±0.63 | 91.98±3.69 | 90.96±5.45 | 99.40±1.58 | 99.19±0.82 | 96.07±3.90 | **99.53±1.08** | 99.19±1.63 |
| Bitumen | 93.11±1.56 | 99.39±0.68 | 88.49±12.03 | 93.80±8.83 | **100.0±0.00** | 99.08±0.76 | 99.58±0.67 | 99.34±0.67 | 99.93±0.20 |
| Bricks | 83.86±3.96 | 98.94±0.80 | 84.79±7.36 | 89.83±6.47 | **98.99±1.43** | 91.74±3.98 | 86.63±7.99 | 97.10±2.75 | 98.50±0.95 |
| Shadow | **99.85±0.12** | 96.66±1.33 | 99.41±0.94 | 96.82±1.69 | 99.47±0.93 | 95.49±2.30 | 95.54±2.17 | 93.95±3.14 | 99.96±0.05 |
| OA (%) | 84.53±2.22 | 95.26±1.74 | 94.72±1.17 | 95.87±1.85 | 95.79±2.55 | 94.06±1.29 | 92.61±1.92 | 94.33±2.85 | **96.40±2.27** |
| AA (%) | 88.63±1.57 | 96.99±0.74 | 93.50±2.07 | 94.75±2.36 | 98.25±0.83 | 94.81±0.54 | 93.37±1.24 | 94.81±1.77 | **98.43±0.77** |
| K×100 | 80.00±2.76 | 93.80±2.23 | 93.02±1.53 | 94.58±2.41 | 94.54±3.24 | 92.22±1.63 | 90.29±2.47 | 92.60±3.63 | **95.31±2.92** |

Table 12. Testing data classification results (mean ± standard deviation) on the Houston dataset

| Class | EMP-SVM | CNN | SSRN | DBDA | MSSG | LSFAT | SSFTT | CT-Mixer | **SS-Mamba** |
|---|---|---|---|---|---|---|---|---|---|
| Grass-healthy | 92.99±4.30 | 92.59±4.56 | **96.25±2.94** | 93.48±5.59 | 93.44±4.63 | 92.67±4.33 | 95.36±3.62 | 90.90±5.13 | 92.88±4.33 |
| Grass-stressed | 93.06±5.72 | 97.00±2.21 | 97.65±2.48 | 95.10±3.78 | 95.37±2.36 | 97.57±1.98 | **98.58±1.13** | 95.11±2.60 | 95.99±2.91 |
| Grass-synthetic | 98.97±1.10 | 98.66±1.41 | 99.93±0.22 | **100.0±0.00** | 98.67±1.26 | 98.17±1.20 | 99.20±0.60 | 97.10±4.20 | **100.0±0.00** |
| Tree | 94.75±2.94 | 97.66±1.81 | 95.98±4.13 | 97.13±2.17 | 97.69±1.74 | 96.58±1.78 | 97.82±3.19 | 92.27±4.20 | **99.17±1.67** |
| Soil | 96.51±4.52 | 97.47±5.11 | 95.41±2.36 | 97.66±2.42 | 98.61±3.22 | 99.78±0.29 | **99.99±0.02** | 98.58±3.08 | 98.29±5.08 |
| Water | 94.72±3.42 | 95.38±3.52 | 97.31±7.83 | 97.37±2.23 | 95.08±3.64 | 97.47±3.66 | **98.42±3.86** | 96.42±3.53 | 96.39±3.63 |
| Residential | 85.54±4.67 | 90.54±2.41 | **92.10±2.47** | 91.93±3.62 | 91.46±3.66 | 86.43±2.43 | 84.32±5.32 | 89.18±3.91 | 91.60±2.06 |
| Commercial | 69.36±4.90 | 78.48±6.64 | 93.23±3.49 | **94.88±3.19** | 78.02±5.81 | 83.25±6.19 | 80.80±5.19 | 83.90±4.83 | 83.33±4.35 |
| Road | 75.81±6.81 | 90.60±3.94 | 89.87±3.83 | 88.57±2.27 | 92.26±3.16 | 82.61±5.52 | 78.82±6.79 | 86.58±2.31 | **92.05±2.02** |
| Highway | 87.63±4.01 | 96.06±4.12 | 86.49±6.78 | 89.76±3.83 | 97.61±2.10 | 96.50±2.86 | 95.10±3.33 | **99.09±1.52** | 98.05±1.94 |
| Railway | 85.58±7.72 | 91.52±4.71 | 90.45±1.94 | 95.44±2.10 | 93.82±5.63 | 93.46±5.38 | **95.82±3.23** | 92.34±6.17 | 92.48±6.14 |
| Parking-lot-1 | 76.18±6.14 | 91.81±5.48 | 89.91±4.73 | **93.15±3.27** | 92.45±6.26 | 89.55±5.62 | 89.65±4.87 | 91.27±5.74 | 91.24±5.57 |
| Parking-lot-2 | 56.44±5.90 | 96.08±2.62 | 93.52±5.65 | 82.75±6.06 | 95.86±2.48 | 92.98±4.01 | **97.75±2.10** | 91.22±5.39 | 92.87±3.30 |
| Tennis-court | 97.94±2.53 | 99.93±0.22 | 97.67±3.29 | 98.13±2.59 | 99.95±0.15 | 99.88±0.25 | 99.85±0.22 | **100.0±0.00** | **100.0±0.00** |
| Running-track | 99.08±0.46 | 99.59±1.06 | 96.93±1.96 | 95.05±2.47 | 99.45±1.59 | 99.97±0.09 | **100.0±0.00** | 98.63±2.38 | **100.0±0.00** |
| OA (%) | 86.56±1.36 | 93.36±0.92 | 93.32±1.05 | 93.67±0.92 | 93.92±1.02 | 92.85±0.93 | 92.88±0.97 | 92.73±1.21 | **94.30±1.10** |
| AA (%) | 86.97±1.26 | 94.23±0.67 | 94.18±1.12 | 94.03±0.87 | 94.65±0.85 | 93.79±0.74 | 94.10±0.85 | 93.51±1.04 | **94.96±0.89** |
| K×100 | 85.47±1.47 | 92.82±0.99 | 92.78±1.13 | 93.16±1.00 | 93.43±1.10 | 92.27±1.01 | 92.30±1.05 | 92.14±1.31 | **93.84±1.20** |

Table 13. Testing data classification results (mean ± standard deviation) on the Chikusei dataset

| Class | EMP-SVM | CNN | SSRN | DBDA | MSSG | LSFAT | SSFTT | CT-Mixer | **SS-Mamba** |
|---|---|---|---|---|---|---|---|---|---|
| Water | 83.55±10.60 | 92.99±4.40 | 83.51±12.94 | 83.44±13.8 | 94.58±4.95 | 94.77±4.82 | 93.86±6.35 | 90.75±5.96 | **96.00±2.62** |
| Bare soil (school) | 93.83±3.84 | 99.54±0.53 | 98.07±2.02 | 99.65±0.51 | 99.72±0.32 | 99.73±0.29 | 99.38±0.53 | 99.76±0.39 | **100.0±0.00** |
| Bare soil (park) | 98.01±2.62 | 99.57±0.98 | 28.93±10.75 | 31.63±15.77 | **100.0±0.00** | 99.50±0.99 | 99.03±1.70 | 99.89±0.32 | 99.72±0.85 |
| Bare soil (farmland) | 50.19±20.7 | 82.66±16.10 | **90.14±11.38** | 87.22±10.73 | 83.77±14.7 | 81.21±15.6 | 82.93±16.13 | 83.00±18.13 | 84.44±17.5 |
| Natural plants | 96.70±2.76 | 99.95±0.02 | 95.10±3.32 | 96.53±3.24 | 99.99±0.02 | **100.0±0.00** | 99.97±0.05 | 99.49±0.59 | 99.98±0.03 |
| Weeds | 87.28±12.13 | 95.62±3.64 | 73.53±22.89 | 85.41±24.14 | 95.69±3.72 | 94.89±3.55 | 95.17±3.75 | **95.26±3.85** | 95.01±3.60 |
| Forest | 82.13±7.49 | 99.97±0.05 | 95.66±3.70 | 99.37±0.87 | 99.96±0.03 | 99.67±0.54 | 96.73±9.24 | 99.92±0.07 | **100.0±0.00** |
| Grass | 91.93±2.72 | 93.05±2.99 | 96.71±4.96 | **99.90±0.27** | 92.91±3.52 | 91.95±2.20 | 92.40±3.60 | 93.78±3.25 | 97.42±3.12 |
| Rice field (grown) | 79.34±20.97 | 94.59±10.58 | 96.57±3.77 | **99.43±0.46** | 91.72±12.0 | 93.43±1.12 | 88.42±13.33 | 86.40±13.20 | 94.08±11.0 |
| Rice field (first stage) | 99.26±0.55 | 99.94±0.17 | 81.93±9.86 | 89.73±5.23 | **100.0±0.00** | **100.0±0.00** | **100.0±0.00** | **100.0±0.00** | **100.0±0.00** |
| Row crops | 66.40±14.47 | 82.22±10.90 | 94.58±11.37 | **97.42±3.29** | 83.82±11.0 | 82.52±10.40 | 83.43±14.00 | 85.10±10.49 | 83.96±9.60 |
| Plastic house | 69.20±11.50 | 84.48±9.13 | 91.50±6.20 | **96.78±4.48** | 86.69±8.30 | 68.08±22.0 | 79.03±12.18 | 90.79±10.63 | 85.37±8.23 |
| Manmade-1 | 95.09±1.97 | 95.97±1.48 | 96.16±7.62 | 98.75±2.27 | 96.22±1.66 | 96.93±1.12 | **97.15±1.79** | 96.44±1.78 | 96.36±1.62 |
| Manmade-2 | 86.85±11.24 | 89.49±10.80 | **99.80±0.33** | 99.60±7.82 | 94.45±6.96 | 92.73±9.58 | 91.19±10.60 | 93.04±8.38 | 92.95±7.95 |
| Manmade-3 | 91.01±17.23 | 91.78±8.43 | 93.87±9.69 | **98.12±5.20** | 97.96±4.20 | 96.31±10.98 | 93.73±12.62 | 95.59±10.38 | 93.97±130 |
| Manmade-4 | 93.73±7.85 | 95.67±6.04 | 93.60±7.32 | **98.24±3.48** | 94.70±7.96 | 95.48±8.05 | 94.19±4.49 | 94.52±8.03 | 97.33±7.71 |
| Manmade grass | 93.39±6.38 | 96.06±8.39 | 98.35±1.65 | 96.62±2.32 | 99.71±3.46 | 98.40±4.12 | 99.74±0.75 | **100.0±0.00** | **100.0±0.00** |
| Asphalt | **88.52±12.17** | 83.98±11.2 | 69.53±13.85 | 72.33±13.82 | 85.43±12.20 | 79.53±18.73 | 78.37±19.95 | 76.36±17.46 | 85.14±13.50 |
| Paved ground | 88.07±7.69 | 98.86±3.43 | 24.50±16.27 | 35.81±35.81 | **100.0±0.00** | 99.93±0.21 | 99.29±0.95 | **100.0±0.00** | **100.0±0.00** |
| OA (%) | 81.58±4.64 | 93.87±2.28 | 91.46±3.62 | 94.39±2.39 | 94.28±2.64 | 93.37±2.61 | 92.05±3.26 | 93.17±3.25 | **94.97±2.34** |
| AA (%) | 86.02±3.06 | 93.50±1.40 | 84.32±2.98 | 88.39±2.24 | 94.59±1.63 | 92.90±1.73 | 92.84±1.71 | 93.69±2.14 | **94.83±1.58** |
| K×100 | 78.97±5.31 | 92.95±2.60 | 90.18±4.13 | 93.55±2.73 | 93.43±3.01 | 92.39±2.97 | 90.87±3.70 | 92.16±3.69 | **94.22±2.67** |

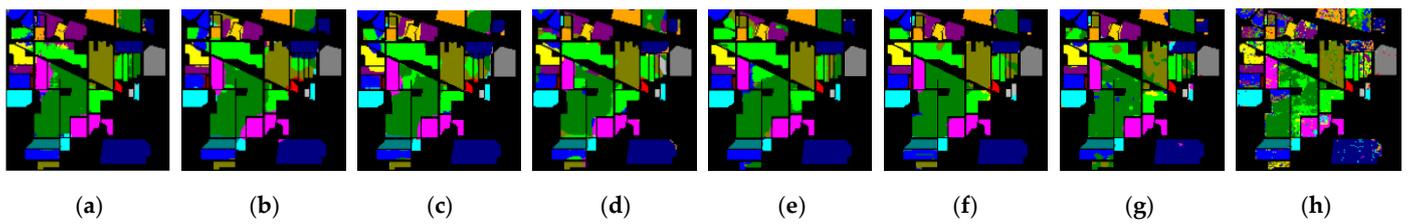

**Figure 8.** Classification maps using different methods on the Indian Pines dataset. (a) SS-Mamba, (b) CT-Mixer, (c) SSFTT, (d) LSFAT; (e) MSSG, (f) DBDA, (g) SSRN; (h) EMP-SVM.

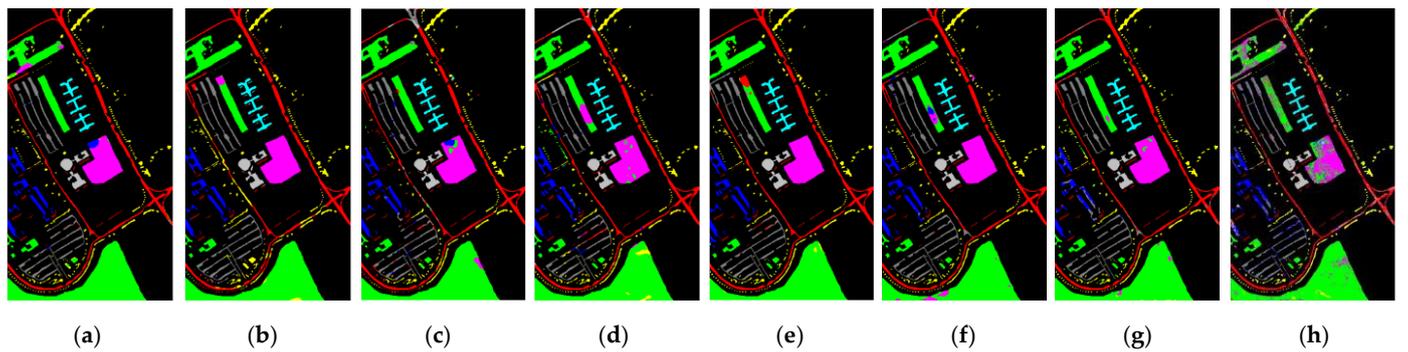

**Figure 9.** Classification maps using different methods on the Pavia University dataset. (a) SS-Mamba, (b) CT-Mixer, (c) SSFTT, (d) LSFAT; (e) MSSG, (f) DBDA, (g) SSRN; (h) EMP-SVM.

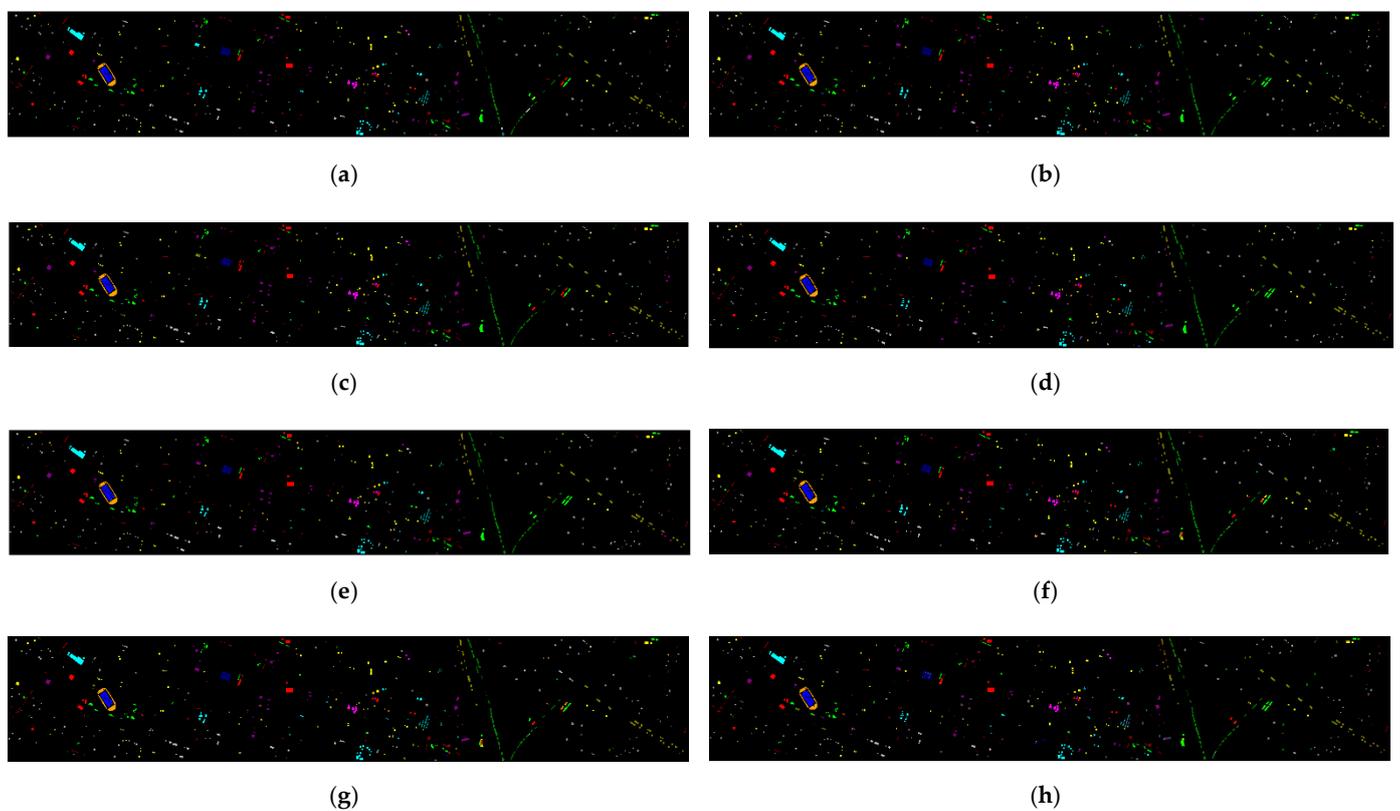

**Figure 10.** Classification maps using different methods on the Houston dataset. (a) SS-Mamba, (b) CT-Mixer, (c) SSFTT, (d) LSFAT; (e) MSSG, (f) DBDA, (g) SSRN; (h) EMP-SVM.

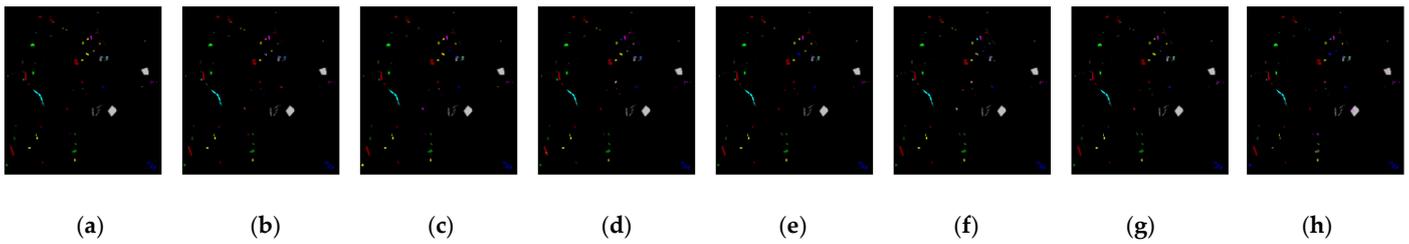

(**a**)　　　(**b**)　　　(**c**)　　　(**d**)　　　(**e**)　　　(**f**)　　　(**g**)　　　(**h**)

**Figure 11.** Classification maps using different methods on the Chikusei dataset. (a) SS-Mamba, (b) CT-Mixer, (c) SSFTT, (d) LSFAT; (e) MSSG, (f) DBDA, (g) SSRN; (h) EMP-SVM.

*3.5. Classification Maps*

As qualitative classification results, the classification maps for different methods on the four datasets are shown in Figures. 8-11. From these maps, it can be clearly seen that the proposed SS-Mamba achieves better classification performance than the comparison methods. For example, on the Pavia dataset, many methods tend to misclassify some pixels from *Asphalt, Trees* and *Bricks*. The reason may be that they are located close to each other. This suggests that these models may focus more on spatial information brought by spatial window, potentially neglecting the spectral characteristics for accurate classification. In contrast, the proposed method demonstrates good performance in this context. Thanks to the designed spectral-spatial learning framework and the feature extraction ability of Mamba, it can make full use of spectral–spatial information and improve the performance of HSI classification.

**4. Discussion**

*4.1. Complexity Analysis*

**Table 14**. Complexity analysis for different models

|  | CT-Mixer | SS-LSTM | SS-GRU | SS-Transformer | SS-Mamba |
|---|---|---|---|---|---|
| Param. | 0.77M | 1.00M | 0.81M | 0.48M | 0.47M |
| Test Time | 8.61ms | 7.77ms | 9.14ms | 11.05ms | 10.45ms |
| OA (%) | 94.33 | 95.51 | 95.95 | 94.99 | 96.40 |

Table 14 shows the number of parameters (i.e., Param.) and consuming time (i.e., Test Time) when tested on a batch of 100 samples on the Pavia University dataset. The SS-LSTM/GRU/Transformer means the corresponding spectral-spatial sequence model in the ablation experiment.

The results reveals that the Mamba model exhibits faster inference times compared to the Transformer model. Additionally, sequential models like Mamba generally necessitate longer inference times but also achieve higher classification accuracies when compared to CNN models. By the way, the newly proposed Mamba-2 (i.e., the second generation of Mamba) has faster training and inference process, which will further benefit our work in the future.

*4.2. Features Maps*

Taking the two input samples of the Pavia University as an example, Figure 12 shows the corresponding spatial feature maps of each block. And different columns show the feature maps on different feature channels (i.e., the first 8 channels). It can be seen that the model still retains some of the image details in the token of each block. On the one hand, different channels focus on different image information and their feature maps are not the same. On the other hand, the feature maps of different blocks have some differences. And the image details (semantic structure) are gradually blurred with the depth. However, probably because the model is shallow with a few blocks, the change of feature maps with depth is not particularly obvious.

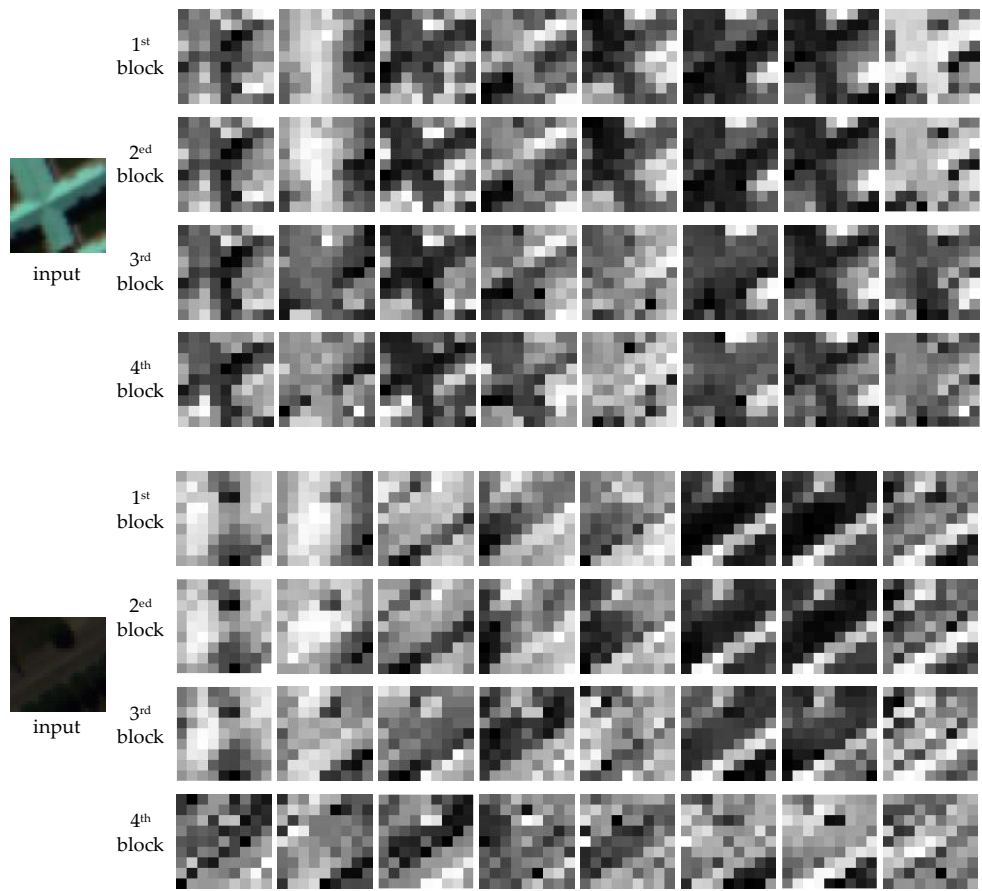

**Figure 12.** The spatial feature maps of each block. For simplicity of illustration, only the spatial feature maps on the first 8 channels are given. Each row corresponds to the feature maps of different channels, while each column represents the feature maps outputted by different blocks.

*4.3. Comparison of Proposed Classification Method with Spectral Unmixing-based Methods*

Spectral unmixing technique plays an important role in HSI processing and application and there are many classification methods which utilize spectral unmixing for improving performance. Therefore, it is necessary to discuss the proposed SS-Mamba's advantages over the spectral unmixing-based classification methods. In [57], hard examples were firstly selected based on specific criteria. Then spectral unmixing was used to improve their quality, facilitating subsequent classification tasks. In [58], the researchers focused on designing autoencoders for unmixing tasks based on the linear mixing model and then using the encoder's features for classification. Li et al. collected the abundance representations from multiple HSIs to form an enlarged dataset, and the used the enlarged abundance dataset to train a classifier like CNN, which could alleviate the overfitting problem [59]. Based on these relevant works, we would like to emphasize the distinctive advantages of our classification approach over these spectral unmixing-based methods.

While spectral unmixing methods are effective in enhancing image quality and extracting spectral information, the whole classification methods typically involve a multi-step process that includes spectral unmixing followed by separate feature extraction and classification steps. This can be time-consuming and requires prior knowledge of hyperspectral mixing models, which may vary in accuracy and impact the final classification results.

In contrast, our proposed classification method is designed as an end-to-end framework. This streamlined approach eliminates the need for intermediate processing stages and expert knowledge of spectral unmixing models, making it more efficient and user-friendly. By leveraging the capabilities of Mamba for feature extraction and classification simultaneously, our method offers a more seamless and accessible solution for achieving high classification accuracy.In the feature, we can try to combine the proposed method and rspectral unmixing technique to obtain better classification performance.

*4.4. Limitations and Feature Work*

The SS-Mamba divides the HSI samples as tokens, which are the input of the SS-Mamba. The token generation process may disrupt the semantic structure of the HSI samples to some extent. Specifically, the current partitioning method does not take object's orientation and shape into consideration, potentially causing pixels belonging to the same object to be distributed across different patches. Therefore, the future work will focus on the combination with traditional architectures like CNN for enhancing local feature extraction and improve the generation of tokens.

## 5. Conclusions

In this study, we make an effective exploration to build a spectral-spatial model for HSI classification, which purely using an emerging sequence model named Mamba. The proposed model converts any given HSI cube to spatial and spectral tokens as sequences. And then stacked Mamba blocks are used to effectively model the relationships between the tokens. Besides, the spectral and spatial tokens are enhanced and fused for more discriminant features. The proposed SS-Mamba is then evaluated on four widely used datasets (i.e., Indian Pines, Pavia University, Houston and Chikusei datasets), and some main conclusions can be drawn from the experimental results as follows:

1) Through a comparative analysis of classification results, it is evident that the proposed SS-Mamba can make full use of spatial-spectral information, and it can achieve superior performance for HSI classification task.
2) The ablation experiments show that as a sequence model, Mamba is effective and can gain competitive classification performance for HSI classification when compared with other sequence models like Transformer, LSTM, and GRU.
3) The ablation experiments also show that the designed spectral-spatial learning framework is effective for different sequence models, when compared with spectral-only or spatial-only models.
4) The designed feature enhancement module is effective to enhance spectral and spatial features and improve the SS-Mamba's classification performance.

This research explored the prospects of the utilization Mamba model in HSI classification and provides new insights for further studies.

**Author Contributions:** Conceptualization: Y.C.; methodology: L.H. and Y.C.; writing—original draft preparation: L.H., Y.C. and X.H. All authors have read and agreed to the published version of the manuscript.

**Funding:** This research was funded by the Natural Science Foundation of China Grants 62371169, 61971164 and U20B2041.

**Institutional Review Board Statement:** Not applicable

**Informed Consent Statement:** Not applicable

**Data Availability Statement:** The Houston dataset is available at: https://hyperspectral.ee.uh.edu/. The Indian Pines and Pavia University datasets are available at: http://www.ehu.eus/ccwintco/index.php?title=Hyperspectral_Remote_Sensing_Scenes (accessed on 1 September 2020). The Chikusei dataset is available at: https://naotoyokoya.com/Download.html (accessed on 1 September 2020).

**Conflicts of Interest:** The authors declare no conflict of interest.